\pdfoutput=1

\documentclass[11pt]{article}

\usepackage[]{acl/ACL2023}

\usepackage{times}
\usepackage{latexsym}

\usepackage[T1]{fontenc}

\usepackage[utf8]{inputenc}

\usepackage{microtype}

\usepackage{inconsolata}

\usepackage{booktabs, multirow, tabularx, csquotes, amsmath, amsthm, amssymb, lipsum, textcomp, pdfpages, pifont, xcolor, caption, subcaption, adjustbox, paralist, hyperref}

\newcommand{\Dataname}{\textsc{kitmus}}

\newcommand{\bgparam}{\textsc{Background-Pretrain}}
\newcommand{\bgparamcont}{\textsc{Background-Both}}
\newcommand{\bgcont}{\textsc{Background-Inference}}
\newcommand{\para}[1]{\smallskip\noindent\textbf{#1}}

\title{The \textsc{KITMUS} Test: Evaluating Knowledge Integration from \\ Multiple Sources
}

\author{
    Akshatha Arodi\textsuperscript{\textnormal 1}\textsuperscript{\textnormal 2}\thanks{\quad Equal contribution}, 
    Martin Pömsl\textsuperscript{\textnormal 1}\textsuperscript{\textnormal 2}\footnotemark[\value{footnote}], 
    Kaheer Suleman\textsuperscript{\textnormal 3}, \\
    {\bf Adam Trischler\textsuperscript{\textnormal 3}, 
    Alexandra Olteanu\textsuperscript{\textnormal 3}, 
    Jackie Chi Kit Cheung\textsuperscript{\textnormal 1}\textsuperscript{\textnormal 2}}\\
    \textsuperscript{1} McGill University \quad \textsuperscript{2} Mila Quebec AI Institute \quad \textsuperscript{3} Microsoft Research \\
    \texttt{\{akshatha.arodi@mail, martin.pomsl@mail, jcheung@cs\}.mcgill.ca}
    \\
    \texttt{\{kaheer.s\}@gmail.com \quad \{adam.trischler, alexandra.olteanu\}@microsoft.com} \\
}

\begin{document}

\maketitle

\begin{abstract}

    Many state-of-the-art natural language understanding (NLU) models are based on pretrained neural language models. These models often make inferences using information from multiple sources.
    %
    An important class of such inferences are those that require both background knowledge, presumably contained in a model's pretrained parameters, and instance-specific information that is supplied at inference time.
    %
    However, the integration and reasoning abilities of NLU models in the presence of multiple knowledge sources have been largely understudied.
    %
    In this work, we propose a test suite of coreference resolution subtasks that require reasoning over multiple facts. These subtasks differ in terms of which knowledge sources contain the relevant facts. We also introduce subtasks where knowledge is present only at inference time using fictional knowledge.
    %
    We evaluate state-of-the-art coreference resolution models on our dataset.
    %
    %
    Our results indicate that several models struggle to reason on-the-fly over knowledge observed both at pretrain time and at inference time. However, with task-specific training, a subset of models demonstrates the ability to integrate certain knowledge types from multiple sources. Still, even the best performing models seem to have difficulties with reliably integrating knowledge presented only at inference time.

\end{abstract}

\section{Introduction}

Progress on natural language understanding (NLU) benchmarks has recently been driven by pretrained large language models (LLMs), which can be adapted to specific tasks via finetuning \citep{peters-etal-2018-deep, devlin-etal-2019-bert, le-scao-rush-2021-many}. These models draw on a variety of knowledge sources, such as knowledge given in inputs at inference time and knowledge contained in their parameters, usually acquired via pretraining.

Recent work suggests that models can use pretrain-time knowledge in tasks like translation and question answering to obtain performance gains \citep{brown2020fewshot, roberts-etal-2020-much}. However, natural language understanding often requires knowledge that is only supplied at inference time because of, e.g., time sensitivity or instance specificity. Consider the passage \enquote{John saw the newly elected president on TV.} Pretrained parameters can conceivably contain information about what presidents do and what a TV is, but they cannot contain reliable knowledge about who John is (since \enquote{John} is an instance-specific identifier) or who the president is (because the president might have changed since pretraining). It follows that successful models for knowledge-intensive NLU tasks might require the ability to use both pretrain-time and inference-time knowledge.

\begin{figure}[t!]
\centering \footnotesize
\begin{tabular}{p{0.95\linewidth}} 
1. \textcolor{brown}{\textbf{Servin}} is a judge. \textcolor{blue}{\textbf{Kea}} is a baker. Servin and Kea met at a park. After a long day at work deciding cases in a law court, \textcolor{red}{\textbf{he}} was happy to relax.~[Answer: \textcolor{brown}{\textbf{Servin}}]
\\
\vspace{0.05mm}
2. Schwing is a gladiower. The work of a gladiower is inwaging ledmonly. The work of a popesmer is chodoling larely. Bate is a popesmer. At the coffee shop, \textcolor{brown}{\textbf{Schwing}} and \textcolor{blue}{\textbf{Bate}} connected. The coffee was excellent. \textcolor{red}{\textbf{She}} shared experiences from a career of chodoling larely.\\\noindent [Answer: \textcolor{blue}{\textbf{Bate}}]
\end{tabular}
\caption{Examples from {\Dataname} showing coreference cases that require real (1.) and fictional (2.) knowledge. To resolve the pronoun (in red), a model needs to draw on entity-specific knowledge about an entity's occupation as well as on background knowledge about what kind of work the occupation entails. 
}
\label{table:example-task}
\end{figure}

To effectively use these two knowledge sources, models must (1) retrieve relevant information from each knowledge source, (2) adjudicate between potentially conflicting information, and (3) integrate multiple units of information from both knowledge sources and reason over them on the fly. For example, pretrained parameters might contain the knowledge that Donald Trump is the president of the United States, but inference-time inputs might state that Joe Biden is the president. Based on the contextual information available in a task, models must infer the correct president.

Studying whether current models can use multiple knowledge sources effectively can help identify and debug errors that models make when relying on outdated sources. To this end, we introduce a coreference resolution task designed to probe models' ability to draw on knowledge available in different sources. Recent work by \citet{longpre-etal-2021-entity} examined the effects of knowledge conflicts across different knowledge sources. In our work, we aim to more broadly examine the behaviour of NLU models in the presence of multiple knowledge sources. While \citet{longpre-etal-2021-entity} study how models handle conflicting facts, our goal is to evaluate whether models can combine complementary knowledge drawn from multiple sources rather than choose between sources.

In our task, the resolution of a given pronoun requires two types of knowledge (see Figure~\ref{table:example-task}):
\begin{inparaenum}[1)]
\item entity-specific knowledge, e.g., \enquote{Servin is a judge} and 
\item background knowledge, e.g., \enquote{Judges decide cases in law courts.}
\end{inparaenum}
Generally, background knowledge is learned during the pretraining of LLMs i.e., at pretrain-time, while entity-specific knowledge is typically observed at inference time. We vary the availability of the required information such that it may either be found in a single source or in multiple sources. We evaluate a model's ability to integrate and reason over the two knowledge types (entity-specific and background knowledge), given in two knowledge sources (pretrain-time and inference-time).

We propose {\Dataname},\footnote{\url{https://github.com/mpoemsl/kitmus}} a diagnostic test suite. 
The {\Dataname} tests evaluate \textsc{k}nowledge \textsc{i}n\textsc{t}egration from \textsc{mu}ltiple \textsc{s}ources. {\Dataname}'s distinguishing feature is that it contains texts in which we methodically vary the mapping of knowledge types to knowledge sources, which allows us to pinpoint the specific strengths and limitations of models. We also analyze the behaviour of models when knowledge is available only at inference-time by introducing variants where a model needs to reason over fictional knowledge, which is presumably not contained in the parameters. Unlike previous reasoning datasets, where inference-time knowledge is retrieved \citep{onoe2021creak}, we provide the knowledge necessary to solve the task in each instance of {\Dataname}. This allows for a more controlled setting where we can focus on knowledge integration, rather than on retrieval, which we hold out as a separate problem. In a study with human participants, we empirically validated that both entity-specific and background knowledge are required to perform well on {\Dataname}, and that the automatically generated labels are consistent with human annotations.

We evaluate state-of-the-art coreference resolution models on the {\Dataname}. In our experiments, many established models appear unable to integrate knowledge from two different knowledge sources and reason over them. With task-specific training, two models---BERT4Coref \citep{joshi-etal-2019-bert} and C2F \citep{lee-etal-2018-higher}---demonstrate the ability to reason over both knowledge observed at pretrain time and at inference time. However, we find that the ability to integrate knowledge from different sources seems to depend on the knowledge type in that source. While knowledge integration through concatenation at inference time seems to be effective for entity-specific knowledge, experiments with fictional knowledge indicate that even the best performing models cannot reliably integrate all types of background knowledge when provided only at inference time.

\section{Related Work}

\noindent \textbf{Coreference resolution as a reasoning task:} 
There has been extensive work to study NLU models' ability to exploit linguistic knowledge that involves shallow cues such as gender, position, and number cues \citep{durrett-klein-2013-easy}, as well as other properties like semantic roles \citep{baker-etal-1998-berkeley-framenet, chambers-jurafsky-2009-unsupervised}. The Winograd Schema Challenge (WSC) \citep{levesque2012winograd} inspired a number of specialized datasets such as GAP \citep{webster-etal-2018-mind} and Winogrande \citep{sakaguchi2020winogrande} where coreference resolution is used as a test bed for reasoning over knowledge and cases cannot be solved with shallow features \citep{emami-etal-2019-knowref, rahman-ng-2012-resolving}. Following this line of work, we use templates that omit shallow cues, such that a model must integrate knowledge about the world to determine the coreference. While WSC and KnowRef focus on abstract external knowledge that is valid independent of the specific entities involved \citep{emami-etal-2019-knowref}, {\Dataname} is more diverse and allows both entity-specific and background knowledge.

\para{World knowledge for reasoning tasks:} Prior work has shown that integrating world knowledge can lead to improvement in coreference solvers. \citet{bean-riloff-2004-unsupervised} learn caseframe co-occurrence statistics, which they use to predict coreference. \citet{rahman-ng-2012-resolving, zhang-etal-2019-knowledge, aralikatte-etal-2019-rewarding, emami-etal-2019-knowref} showed improved results using data augmentation.

\citet{longpre-etal-2021-entity} recognized the distinction between pretrain-time and inference-time knowledge, but they call them parametric and contextual knowledge. In the context of our work, the term \enquote{contextual} has many different interpretations and could consequently lead to misunderstandings. Therefore, we instead focus on the time at which the knowledge is typically observed in order to distinguish the two knowledge sources.

\citet{chan2022transformers} show that transformers exhibit different inductive biases in how they represent and generalize from the information in pretrain-time and inference-time knowledge sources using synthetic sequences of characters. \citet{mallen2022not} 
probe LMs on factual knowledge memorization using open-domain question answering and show improved results with retrieved knowledge augmentation. Complementing prior tasks that require background knowledge found in off-the-shelf knowledge bases, {\Dataname} instances require both entity-specific and background knowledge---we map a mentioned entity to its occupation and occupations to situations. \citet{onoe2021creak} pose fact-checking tasks that require combining entity knowledge with commonsense knowledge. In contrast to our dataset, they do not provide the required knowledge, and expect models to either use only pretrain-time knowledge in a closed-book setting or to retrieve the knowledge from an external knowledge base at inference time. In our work, the knowledge associated with each instance is generated in a controlled way and provided as part of the inputs. 

\para{Reasoning over knowledge with Transformers:} \citet{zhou-etal-2021-rica} present a dataset that evaluates the ability of pretrained Transformer language models to make inferences over axioms stated in natural language. Similarly, \citet{clark2020transformers} study the limits of reasoning in Transformer models with an approach where classical logic facts and rules are stated using natural language instead of a formal representation. Though our task is presented as a natural language text that requires reasoning, and is evaluated on Transformer models (among others), our work differs from \citet{zhou-etal-2021-rica} and \citet{clark2020transformers}'s in that the prediction target is the resolution of pronoun coreferences within a text. This requires identifying those mentions of an entity in a text that corefer with a pronoun using both pretrain-time and inference-time knowledge. In contrast, \citet{zhou-etal-2021-rica} and \citet{clark2020transformers} predict whether a conclusion is consistent with a preceding premise.

\section{The {\Dataname} Test Suite}
\label{sec:dataset}

\begin{figure}[t!]
    \centering
    \includegraphics[scale=0.04]{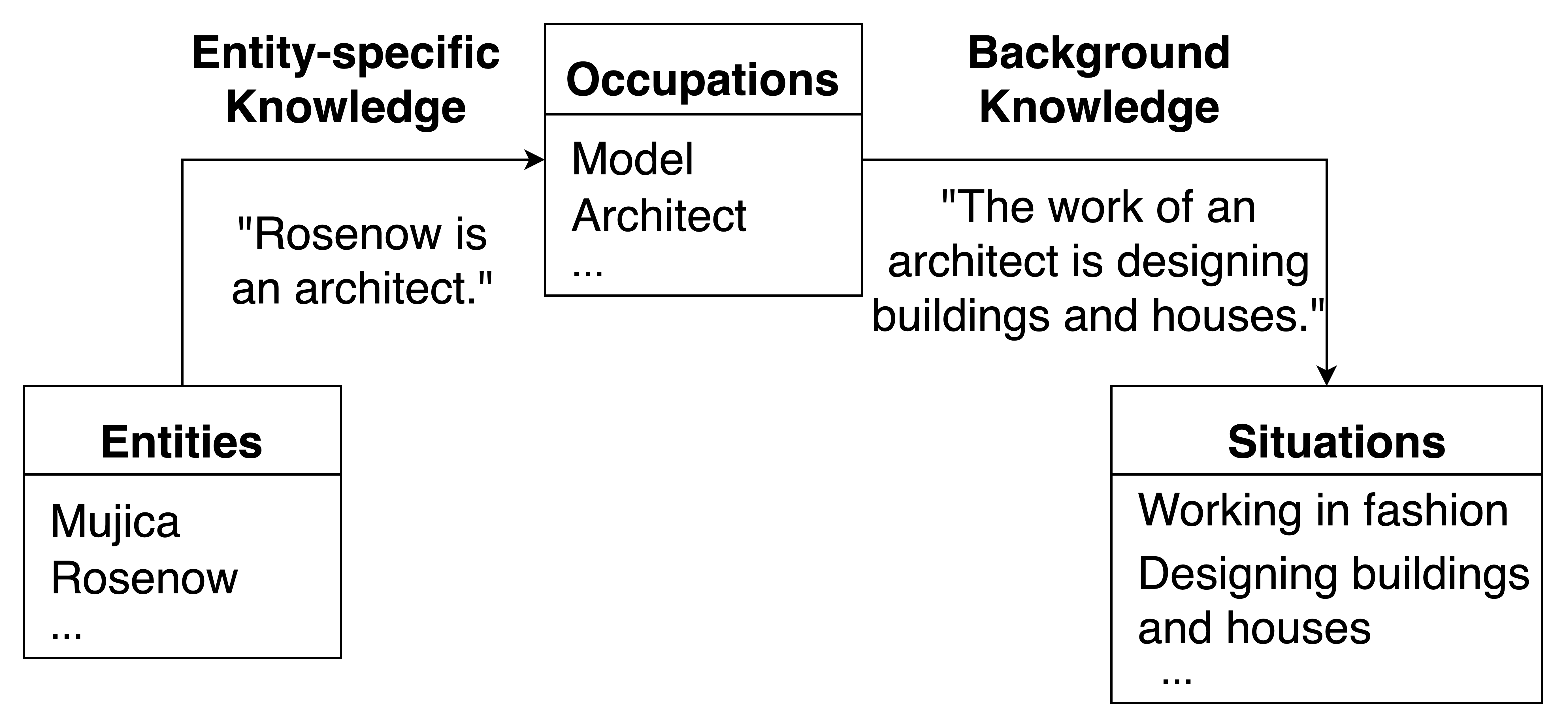}
    \vspace{-8pt}
    \caption{Schema of knowledge types in {\Dataname}.}
    \vspace{-8pt}
    \label{fig:knowledge-schema}
\end{figure}


We evaluate the knowledge integration capability of coreference resolution models from different knowledge sources: \begin{inparaenum}[1)] 
\item pretrain-time: knowledge accumulated in the parameters during (pre-)training and 
\item inference-time: knowledge observed in an input text.
\end{inparaenum}

To design {\Dataname}, we formulate a coreference resolution task which requires access to two facts. We systematically vary the presence of these facts across the knowledge sources to evaluate the models. As an instantiation of the idea of presenting two facts, we experiment with the following two knowledge types:

\begin{compactitem}[--]
    \item {\bf Entity-specific}: occupation of an entity e.g., \enquote{Rosenow is an architect.}
    \item {\bf Background}: situation typical for an occupation e.g., \enquote{architects design building and houses.}
\end{compactitem}

For example, consider the following task to predict whether Mujica or Rosenow is the correct antecedent of the pronoun \enquote{he.}

\begin{quote} \footnotesize
    Mujica is a model. Rosenow is an architect. At the bus station, \textcolor{brown}{\textbf{Mujica}} and \textcolor{blue}{\textbf{Rosenow}} connected. Public transports are eco-friendly. \textcolor{red}{\textbf{He}} shared experiences from a career of designing buildings and houses.~[Answer: \textcolor{blue}{\textbf{Rosenow}}]
\end{quote}

Here, the occupations are \textit{model} and \textit{architect}, and the situational cue is \textit{designing building and houses}. Both knowledge types are required in order to resolve this coreference. An illustration of this knowledge schema can be found in Figure~\ref{fig:knowledge-schema}.

\begin{figure*}[t!]

\centering

\subcaptionbox{\bgparam\label{fghfrgh}}[0.325\textwidth]
{\includegraphics[scale=0.048]{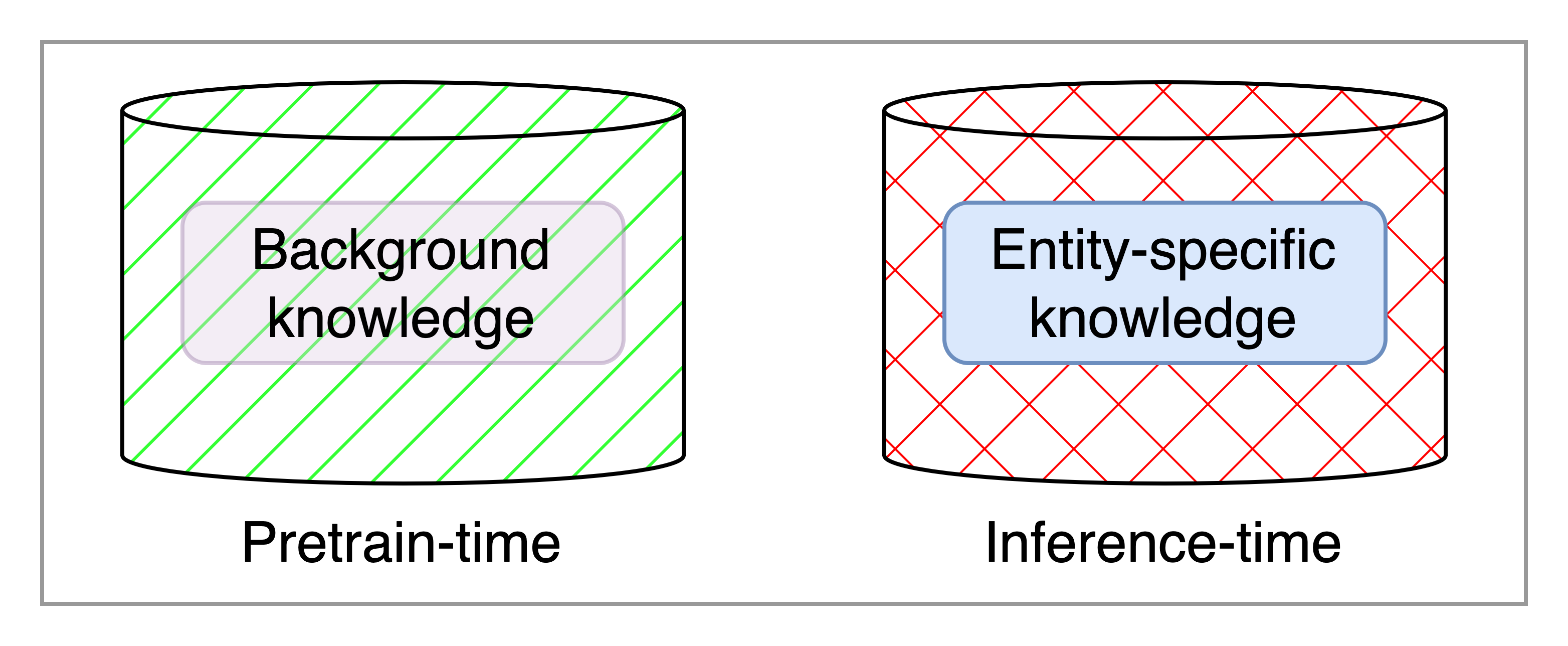}
}
\hspace*{\fill}
\subcaptionbox{\bgparamcont\label{sdfgsdfgh}}[0.325\textwidth]
{\includegraphics[scale=0.048]{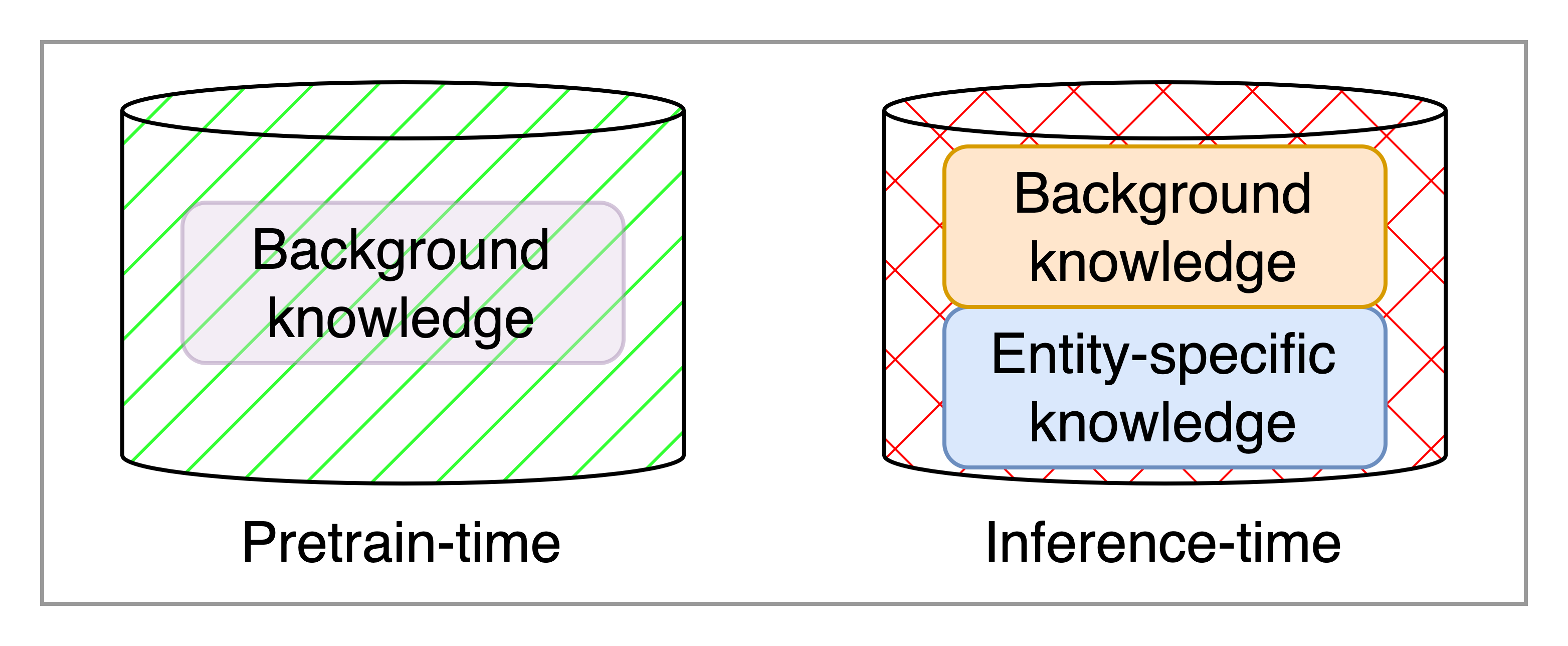}}
\hspace*{\fill}
\subcaptionbox{\bgcont\label{ygvtfvrdc}}[0.325\textwidth]
{\includegraphics[scale=0.048]{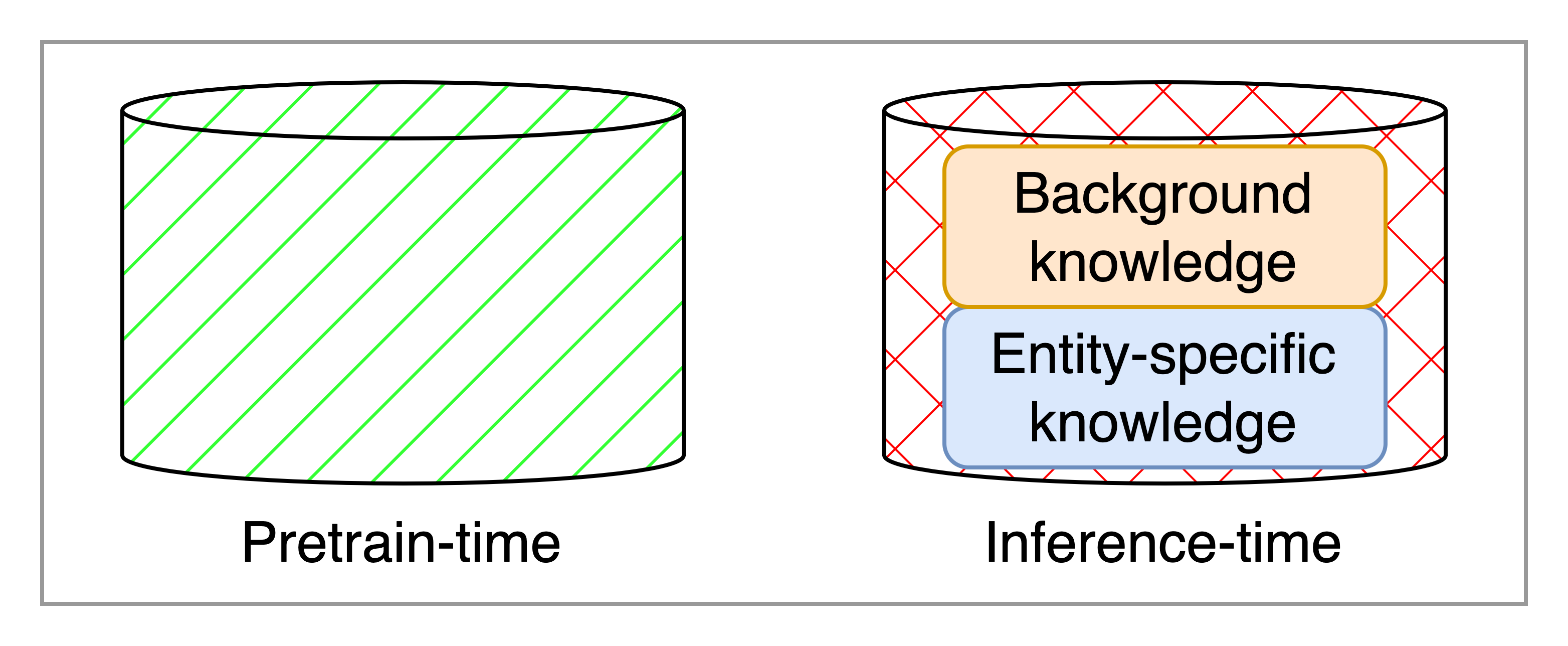}}
\hspace*{\fill}
    \caption{Variants of {\Dataname} based on the source of background knowledge.}
    \label{fig:variants}

\end{figure*}
We explore three main variants of the dataset as shown in Figure \ref{fig:variants}. With entity-specific knowledge always provided in the instance, the variants differ based on when and where background knowledge is available:

\begin{compactenum}[--]
    \item {\bgparam}: Background knowledge is available only in the model parameters
    \item {\bgparamcont}: Background knowledge is available in the model parameters and explicitly provided in the instance
    \item {\bgcont}: Background knowledge is available only in the instance
\end{compactenum}

\noindent Each instance of the {\Dataname} task consists of two fragments of text that are concatenated: 1) a knowledge text---containing the inference-time knowledge that models are given access to---and 2) a task text---consisting of the coreference task that models solve.

\subsection{\bgparam}

In this variant, entity-specific knowledge is provided at inference time and background knowledge about occupations is assumed to be pretrain-time knowledge, since information such as \enquote{architects design buildings and houses} is likely to have been observed during pretraining. An example is shown in the previous section. Here, the entity-specific knowledge about Mujica and Rosenow is inference-time; however, the knowledge about the occupations of a model and architect is pretrain-time. With this variant, we evaluate whether models have the ability to integrate and reason over both pretrain-time and inference-time knowledge effectively.

\subsection{\bgparamcont}

In this variant, background knowledge is provided at both inference-time and assumed to be captured by the parameters. Entity-specific and background facts are present in the same knowledge source. They both represent inference-time knowledge being listed in the knowledge text as part of the inference-time inputs. For example:

\begin{quote} \footnotesize
    \textcolor{brown}{\textbf{Chichester}} is a politician. The work of a politician is seeking an elected seat in government. \textcolor{blue}{\textbf{Klose}} is an astronomer. The work of an astronomer is studying the stars and the universe. \textcolor{brown}{\textbf{Chichester}} and \textcolor{blue}{\textbf{Klose}} met at the train station. After a long day at work seeking an elected seat in government, \textcolor{red}{\textbf{she}} was happy to relax.\\\noindent [Answer: \textcolor{brown}{\textbf{Chichester}}]
\end{quote}

\subsection{\bgcont}

In order to evaluate whether a model can solve this task using exclusively inference-time knowledge (i.e., in the absence of pretrain-time knowledge), we introduce fictional \enquote{knowledge.} Fictional knowledge such as \enquote{the work of a mornisdeiver is gupegaing advaily} is unlikely to have been observed during pretraining, in contrast to real-world knowledge which is likely to have been observed. The entities in all variants are always fictional, ensuring that entity-specific knowledge about them was not observed at pretrain time. Thus, in this variant, both knowledge types are fictional and not contained in the pretrained parameters. For example:

\begin{quote} \footnotesize
    The work of a johumker is toncing ignaftedly. The work of a fangher is sparluing gobewly. Amezcua is a johumker. Hundley is a fangher. \textcolor{brown}{\textbf{Hundley}} and \textcolor{blue}{\textbf{Amezcua}} met at the yoga class. Yoga is best done in silence. \textcolor{red}{\textbf{He}} reflected on whether sparluing gobewly for a living was a good career choice.\\\noindent [Answer: \textcolor{brown}{\textbf{Hundley}}]
\end{quote}

Background knowledge about occupations maps occupations to situations that are typical for the occupation, such as \enquote{astronomer} and \enquote{studying the stars and the universe.} To make background knowledge fictional, either the occupation, the situation, or both have to be fictional. For situations, we furthermore distinguish between levels of fictionality and define two sub-variants: 1) word-level fictional situations that use existing words but describe novel occupations, and 2) character-level fictional situations that use novel words. The methods we use to generate these fictional occupations and situations are detailed in the following section. Example texts resulting from different forms of fictionality can be seen in Table \ref{table:occupation-comb}.

\begin{table*}[t!]
    \centering \footnotesize
    \def\arraystretch{1.6}
    \begin{tabular}{p{0.01\linewidth}  p{0.07\linewidth} p{0.06\linewidth} p{0.76\linewidth}}
    Var. & Occupation & Situation & Example \\
    \midrule
    
   BB & \textcolor{blue}{Real}	& \textcolor{blue}{Real} & The work of a \textcolor{blue}{\textit{politician}} is \textcolor{blue}{\textit{seeking an elected seat in government}}. Chichester is a politician[...]\\
   
   BI & \textcolor{blue}{Real}	& \textcolor{orange}{CharFict} & The work of a \textcolor{blue}{\textit{politician}} is \textcolor{orange}{\textit{ehemting smorbtly}}. Chichester is a politician[...]\\	
   
   BI & \textcolor{blue}{Real}	& \textcolor{olive}{WordFict} &	The work of a \textcolor{blue}{\textit{politician}} is \textcolor{olive}{\textit{controlling the pool of an aircraft by using its directional flight controls}}. Chichester is a politician[...]\\
   
   BI & \textcolor{orange}{CharFict}	& \textcolor{blue}{Real} &	The work of a \textcolor{orange}{\textit{mirituer}} is \textcolor{blue}{\textit{seeking an elected seat in government}}. Chichester is a mirituer[...]\\
   
   BI & \textcolor{orange}{CharFict} &	\textcolor{orange}{CharFict} &	The work of a \textcolor{orange}{\textit{mirituer}} is \textcolor{orange}{\textit{ehemting smorbtly}}. Chichester is a mirituer[...]\\
   
   BI & \textcolor{orange}{CharFict} &	\textcolor{olive}{WordFict} &	The work of a \textcolor{orange}{\textit{mirituer}} is \textcolor{olive}{\textit{controlling the pool of an aircraft by using its directional flight controls}}. Chichester is a mirituer. [...]\\								
    \bottomrule
    \end{tabular}
    \caption{Different combinations of fictional occupations and situations in {\bgcont} (BI) variant. An instance of {\bgparamcont} (BB) variant is also shown.}
    \label{table:occupation-comb}
\end{table*}

\section{Dataset Creation}
\label{sec:generation}


To construct {\Dataname}, we manipulate which entities are mentioned in each instance, what occupations those entities have, what situations those occupations pertain to, what contexts they are mentioned in, and whether noise is present. Each entry is structured to first (1) introduce the entities, (2) then place them in the same location, and (3) finally, place one of them in a situation related to their occupation. In the {\bgparamcont} and {\bgcont} variants, this is preceded by a knowledge text mapping entities to their respective occupations using the phrase \enquote{is a.} 


The dataset entries are generated using hand-crafted English-language templates and sampling from a variety of resource pools to fill the template slots. The use of templates facilitates control over the source a certain type of knowledge is stored in, which may not be possible to do with a natural dataset.

We aim to minimize the likelihood of models learning to exploit any spurious correlations in the templates or resources and promote data diversity using the following methods:

\begin{compactenum}[--]
    \item We use multiple templates for each sentence. Examples are shown in Table \ref{tab:templates} in the Appendix.
    \item We sample from diverse resource pools to fill template slots as detailed in Section \ref{sec:resources}.
    \item We include location-dependent noise statements that act as distractors and serve to vary the distance between entities.
    \item We create canonical train, validation, and test splits for each variant that are generated using disjunct subsets of templates and resources.
\end{compactenum}

With these measures, we ensure that all entity names, occupations, situations, locations, templates, and noise statements that occur in the test instances do not occur in the train instances.

\subsection{Resource Pools}
\label{sec:resources}

We collect $20{,}000$ last names as entities, $60$ common occupations as occupations and their associated job descriptions as situations and $112$ common meet-up places as locations from a mix of governmental and other publicly available resources (see Appendix \ref{sec:appendix-resource-pool} for more details). We manually filter for gender and semantic cues. For example, we remove  the occupations that provide referential gender cues such as \enquote{fire-man} and locations that might provide surface cues related to an entity's occupation.

Pronouns are sampled randomly from both the gendered pronouns \texttt{he} and \texttt{she} as well as gender-indefinite pronouns such as singular \texttt{they} and the neopronouns \texttt{ey} and \texttt{ze} following the gender-inclusive coreference resolution dataset \texttt{GICoref} \citep{cao-daume-iii-2020-toward}. Ideally, we would want the distribution of pronouns to approximate the frequency in naturally occurring text, but few reliable statistics exist to estimate them. We include $40\%$ \texttt{he}, $40\%$ \texttt{she}, $10\%$ \texttt{they}, and $10\%$ neopronouns.

Noise statements are sampled randomly from a collection of statements based on the selected location in order to maintain a natural flow of the text. Each location is associated with $25$ noise sentences. These sentences are generated using GPT-2 \citep{radford2019language}, and then manually verified by the authors not to include cues related to any entity or occupation. 

\subsection{Fictional Occupations}
\label{sec:fict-occs}

To create fictional background knowledge that maps occupations to situations, we create fictional occupations and fictional situations. Following the methodology of \citet{malkin-etal-2021-gpt}, we generate $60$ names of fictional occupation by sampling from a character-level LSTM language model.

\subsection{Dataset Formats}
\label{sec:formats}


Each variant in {\Dataname} consists of three subtasks---based on the number of entities---with increasing difficulty: two entity, three entity, and four entity subtasks. Each subtask has train, validation and test splits with $2000$, $400$, and $2000$ examples respectively. The size of {\Dataname} is comparable to that of the GAP dataset \citep{webster-etal-2018-mind}, which similarly tests for a specific phenomenon in ambiguous pronoun coreference resolution.


We provide the test suite in two different formats which are commonly used by state-of-the-art coreference solvers: the CoNLL 2012 format \citep{pradhan-etal-2012-conll} and the GAP format \citep{webster-etal-2018-mind}. The CoNLL format allows for a comprehensive annotation of mentions of an entity—including in the knowledge text. The GAP format, however, allows for the annotation of only two entities and only one mention per entity.

\subsection{Human Validation}

To investigate whether human assessors agree on the resolution of our test cases and whether this resolution is in agreement with our automatically generated labels, we conduct a human validation study. We also investigate whether our assumption that both background and entity-specific knowledge are required to resolve the cases by including instances where the knowledge text is not provided to human participants.

We created a multiple-choice questionnaire by randomly selecting an instance from each variant of our dataset (e.g., {\bgparam}), from each subtask (e.g., two entities), and from each split (e.g., validation). Additionally, we included one instance from each variant and from each subtask where the participants were only given the task text and not the accompanying knowledge text. A total of $96$ sampled instances were presented to six different participants in random order.

A high inter-annotator agreement of $0.938$ as measured by Fleiss' Kappa \cite{fleiss2003kappa} leads us to believe that human participants agree on the resolution of {\Dataname} test cases. We use accuracy as a measure of agreement with the automatically generated labels and find that mean accuracy aggregated over all participants and subtasks is higher than $0.9$ for all variants when the knowledge text is given. As expected, when neither background nor entity-specific knowledge are given, accuracy is below $0.1$ for all variants, since most participants indicate that the question cannot be answered. This suggests that there are no inadvertent cues that can be exploited by humans to solve the task without having access to the entity-specific knowledge and background knowledge contained in the knowledge text. 

Additional details on collection and processing of resource pools, fictional occupations, dataset formats and human validation are in the Appendix.

\section{Experimental Setup}

\subsection{Model Selection}

In this work, we focus on state-of-the-art and well-known coreference resolution models. We experiment with two families of coreference resolution models: 1) general coreference models and 2) pronoun coreference models.

Models that focus on general coreference resolution are often trained on the large Ontonotes corpus in the CoNLL 2012 format \citep{pradhan-etal-2012-conll}. We include BERT4Coref \citep{joshi-etal-2019-bert} as an example of a state-of-the-art models on CoNLL 2012, C2F \citep{lee-etal-2018-higher}, which is the direct successor to the first end-to-end neural coreference resolution model \citep{lee-etal-2017-end}, and Stanford's statistical \citep{clark-manning-2015-entity} and neural \citep{clark-manning-2016-deep} models.

Models that focus on pronoun coreference resolution are trained on the smaller GAP dataset in the GAP format \citep{webster-etal-2018-mind}. We include GREP \citep{attree-2019-gendered}, the winner of the GAP Kaggle competition and PeTra \citep{toshniwal-etal-2020-petra}, an efficient memory-augmented model.

\subsection{Training}

We conduct task-specific training with all models on the train split of {\Dataname} using their default hyperparameters. The larger general coreference models BERT4Coref and C2F are conventionally not trained on datasets with just $2000$ train instances such as GAP or {\Dataname}, but rather trained on Ontonotes and then evaluated on smaller datasets \citep{joshi-etal-2019-bert}. Since coreference cases in {\Dataname} diverge significantly from those in Ontonotes, we test these models both in the Ontonotes-trained setting and {\Dataname}-trained setting. For these models, we report mean metrics over $6$ runs. We use only the pretrained versions of the Stanford models, since they are conventionally used off-the-shelf. We train the GAP-based models---PeTra and GREP---only on the two entity subtasks following the GAP format constraints outlined earlier. Additional training details are in Appendix \ref{sec:appendix-train-time}.

\begin{table*}[t!]
    \footnotesize
    
    \begin{subtable}[h]{0.45\textwidth}
        \centering
        \begin{tabular}{lccc}
        Model & 2 Entities & 3 Entities & 4 Entities \\
        
        \midrule
    
        BERT4Coref & \textbf{0.43} & \textbf{0.18} & \textbf{0.14} \\
        C2F & 0.34 & \textbf{0.18} & 0.13 \\

        Stanford Neural & 0.20 & 0.10 & 0.09 \\
        Stanford Stat. & 0.05 & 0.02 & 0.01 \\
        
        \midrule
        Random & 0.50 & 0.33 & 0.25 \\
        
        \bottomrule
        \end{tabular}
        \caption{Ontonotes-trained}
        \label{tab:bgparam-ontonotes}
    \end{subtable}
    \hfill
    \begin{subtable}[h]{0.45\textwidth}
        \centering
        \begin{tabular}{lccc}
        Model & 2 Entities & 3 Entities & 4 Entities \\
        
        \midrule
        
        BERT4Coref & \textbf{0.99} & \textbf{0.98} & \textbf{0.94} \\
        C2F & 0.52 & 0.28 & 0.48 \\
        
        GREP\textsuperscript{\textdagger} & 0.49 & - & -\\
        PeTra\textsuperscript{\textdagger} & 0.01 & - & -\\
        
        \midrule
        Random & 0.50 & 0.33 & 0.25 \\
        
        \bottomrule
        \end{tabular}
        \caption{{\Dataname}-trained}
        \label{tab:bgparam-dataname}
    \end{subtable}
    
    \vspace{-8pt}
    \caption{Accuracy on {\bgparam} variant of {\Dataname}. Models marked with {\textdagger} operate on GAP format which only allows for the annotation of two entities. All other models operate on the CoNLL format. F1 scores shown in Table \ref{tab:bgparam-f1} track the accuracy scores. Note that models are not forced to choose between entities.}
    \label{tab:bgparam}
\end{table*}

\subsection{Evaluation}

We evaluate all models on the {\Dataname} test split of each subtask. We use two metrics to assess each model performance: antecedent classification F1 and pronoun accuracy. Antecedent classification F1 is typically used for GAP format datasets. It considers the coreference between each candidate antecedent mention and the pronoun as a binary classification decision i.e., for a text with two entities, it considers two binary predictions and calculates the scores accordingly. Pronoun accuracy considers for each pronoun whether the correct candidate antecedent is predicted by the model, so independent from the number of entities in a text, only one decision is made among all possible candidate antecedents. We compare against a random baseline, which is implemented as random choice among the gold candidate mentions.

\section{Experimental Results}

\subsection{{\bgparam}}

Table \ref{tab:bgparam} shows that none of the evaluated models are able to outperform the random baseline without task-specific training on {\Dataname}. Some models exhibit below random performance, indicating that they may fail to recognize and choose the correct mentions that could be antecedents. When trained on {\Dataname}, BERT4Coref (all) and C2F (for the four-entities subtask) perform significantly better than random, as shown in Table \ref{tab:bgparam-dataname}. The high performance of BERT4Coref and C2F on the {\bgparam} variant suggests that both models have the ability to draw background knowledge from their parameters, entity-specific knowledge from the inference-time inputs, and reason over them on-the-fly with task-specific training.

The performance of all models we experimented with generally decreases as the number of entities increases; which is unsurprising since the more candidate entities there are, the less likely the accidental selection of the correct entity becomes. Moreover, we observe high variance across the six runs of {\Dataname}-trained C2F (see Table \ref{tab:bgparam-5k} in the Appendix \ref{sec:appendix-add-exps}).

In order to explore the effect of noise statements, we conduct additional experiments on the {\bgparam} variant without noise. The removal of noise does not result in a significant performance change (see Table \ref{tab:bgparam-noise} in the Appendix).

\subsection{{\bgparamcont} and {\bgcont}}
 
We conduct additional experiments on the {\bgparamcont} and {\bgcont} variants with BERT4Coref and C2F, since they demonstrate the ability to learn the {\bgparam} variant of the task. In Table \ref{tab:bgothers}, we report results on the four-entity subtask, which Table \ref{tab:bgparam} suggests to be the most challenging. While BERT4Coref's performance on the {\bgparamcont} is comparable to its {\bgparam} variant results, C2F's performance is much worse, suggesting that it cannot effectively absorb the background knowledge provided at inference time and is distracted by it. On the {\bgcont} variant, BERT4Coref seems to be able to integrate background knowledge about fictional occupations by outperforming the random baseline. However, it shows the ability to integrate word-level fictional, but not character-level fictional knowledge.

\begin{table}[t!]
    \centering \footnotesize
    \begin{tabular}{lllcc} 
    Var. & Occupation & Situation & BERT4Coref & C2F \\
    \toprule
    
    BB & \multirow{3}{*}{Real} & Real & 0.96 & 0.09 \\
    
    BI &  & CharFict & 0.25 & 0.18 \\ 
    BI &  & WordFict & 0.48 & 0.08 \\
     
    \midrule
     
    BI & \multirow{3}{*}{CharFict} & Real & 0.43 & 0.08 \\
    BI & & CharFict & 0.26 & 0.18 \\ 
    BI & & WordFict & 0.38 & 0.11 \\
    \bottomrule
    
    \end{tabular}
    \vspace{-8pt}
    \caption{{\Dataname}-trained accuracy on {\bgparamcont} (BB) \& {\bgcont} (BI) variants of {\Dataname} with 4 entities. Random performance is $0.25$.
    }
    \label{tab:bgothers}
\end{table}

\section{Discussion}

\para{Models trained on \enquote{general} coreference datasets fail on {\Dataname}:} The poor performance of Ontonotes-trained models suggests that when trained on general coreference resolution datasets, models learn to exploit surface cues, which does not help when testing on {\Dataname} where such cues are removed. Another factor might be the structure of the texts in {\Dataname}, which are designed to place knowledge in specific knowledge sources. This might affect models' abilities to form useful representations resulting in poor performance of Ontonotes-trained models. These failures suggest that training on (what are meant to be) \enquote{general} datasets is not enough to induce knowledge integration from multiple sources and task-specific training is required.

\para{Effect of dataset format and size:} 
We observe that the models that accept input in the CoNLL format \citep{pradhan-etal-2012-conll} perform better than those models that accept the GAP format \citep{webster-etal-2018-mind}. This indicates that mention annotations in the knowledge text---which only the CoNLL format provides---might be significant.

To evaluate whether the failure cases are due to the small train set size of 2000, we repeat experiments with a train set size of 5000. While we do see some improvements, the general trends persist and our observations remain consistent with the previous results (see the limitations section 
for additional discussion). This suggests that further scaling of the train set size might not be sufficient to improve performance on cases where existing models are currently failing.

\para{Performance of current pretrained LLMs:} BERT4Coref seems to consistently outperform C2F. This might be due to the difference in pretrained LLMs: BERT4Coref uses the Transformer architecture \citep{vaswani2017transformers}, which has been shown to be effective at reasoning tasks presented in natural language form \citep{clark2020transformers} and utilizing information presented in inference-time contexts \citep{petroni2020how}, while C2F uses ELMo \citep{peters-etal-2018-deep}. To verify that BERT and ELMo contain background knowledge mapping occupations to situations, we ran a LAMA probe \citep{petroni2020how}. We find that BERT is more likely to contain the background knowledge compared to ELMo (see Section~\ref{sec:limitations} for details). This corroborates the better performance of BERT on knowledge intensive tasks such as {\Dataname}.

\para{Integration of fictional knowledge:}
As shown in Table \ref{tab:bgothers}, BERT4Coref performs consistently poorly on character-level fictional situations compared to real and word-level fictional situations. An example of character-level fictional occupation knowledge erroneously answered by BERT4Coref is shown below:

\begin{quote}\footnotesize
The work of a remaller is socring clatodemnly. \textcolor{brown}{\textbf{Nims}} is a mamser. \textcolor{blue}{\textbf{Formica}} is a remaller. The work of a mamser is slimbing murstly. At the birthday party, \textcolor{blue}{\textbf{Nims}} and \textcolor{brown}{\textbf{Formica}} ran into each other. The party is filled with local and national celebrities and entertainers. \textcolor{red}{\textbf{She}} shared experiences from a career of socring clatodemnly.\\\noindent [Correct answer: \textcolor{brown}{\textbf{Formica}}; BERT4Coref: \textcolor{blue}{\textbf{Nims}}]
\end{quote}

One possible reason could be BERT's tokenization strategy, which involves pooling subword representations \citep{devlin-etal-2019-bert}. In character-level fictional words, the subwords are meaningless, rendering their representations unhelpful. This is consistent with previous work showing that representations of LLMs for character-level fictional \enquote{Jabberwocky} words are less useful \citep{kasai-frank-2019-jabberwocky} and that the presence of out-of-vocabulary (OOV) words decreases performance of neural models for NLU tasks \citep{schick2020rare, moon2020patchbert, he2021context}.

Despite the character-fictional occupations and situations, we expect the models to resolve the coreferences successfully in this setting. In the given example,
the pronoun \enquote{she} can be resolved by matching the situation \enquote{socring clatodemnly} to the occupation \enquote{remaller} (using the word overlap between the situations and the occupation descriptions) and identifying the correct entity associated with the occupation i.e, Formica.

Humans can successfully make these inferences by matching fictional occupations and situations. However, the current models do not perform better than a random baseline in this setting. Our hope is that eventually, models should be able to handle even knowledge presented in previously unknown terms. Given that languages are forever growing, robustness to neologisms is crucial, considering that OOV words e.g., new occupations like \enquote{TikToker} develop constantly.

\para{Effects of knowledge type:} Experiments on the {\bgparam} variant indicate that BERT4Coref is able to integrate fictional entity-specific knowledge observed at inference time reliably, yet this does not seem to be the case for fictional background knowledge. This suggests that models' ability to integrate and reason over the knowledge on-the-fly depends on the knowledge type---whether the knowledge is background or entity-specific---and not on whether it is fictional or not. One possible explanation could be that LMs observed different frequencies of unseen entities, occupations, and situations during pretraining, which result in a difference in their ability to adapt to novel instances of those categories.

\section{Conclusion}

We investigated the ability of models to integrate knowledge from multiple knowledge sources to resolve linguistic ambiguities in a coreference resolution task. We formulated a task that requires access to two knowledge types, entity-specific and background knowledge, and controlled for two knowledge sources that knowledge is available in, pretrain-time and inference-time.

Our results show that with task- and dataset-specific training, some models have the ability to reason over both knowledge observed at pretrain time and at inference time. For these models, knowledge can be integrated by concatenating textual knowledge to the model inputs. Furthermore, our findings imply that supplying additional information (e.g., from a retriever) at inference time to models can be successful even if the knowledge required for the task has not been observed before. However, in our task this ability seems to require task-specific training and depend on the type of knowledge being supplied.

Future work could explore finetuning models on {\Dataname} to encourage knowledge integration across different sources. One might also consider extending the {\Dataname} test suite to other languages or to create a multilingual test suite. Instructions for using our code and adapting the templates and resources to other languages can be found in Appendix \ref{sec:custom-dataset-creation}.

\section*{Limitations}
\label{sec:limitations}

\textbf{Data diversity:}
As a template-generated dataset, {\Dataname} does not reflect the full diversity of natural data. However, we do not attempt to emulate the diversity of natural datasets. Using templates over natural data for diagnostic purposes has a few advantages. Templates facilitate control over the source of a certain type of knowledge, which may not be possible to do with more natural datasets like Ontonotes. This allows us to isolate the model behavior we want to probe. We also take several steps to add diversity, like using multiple templates, sampling from large resource pools, random shuffling of entities, addition of noise sentences, and canonical data splits with non-overlapping templates and resources. To prevent spurious factors at lexical level, the templates are hand-crafted to remove surface cues and validated in a study with human participants.

\para{Background Knowledge Assumption in LMs:} The results of our work is based on the assumption that pretrained LMs have access to background knowledge about real occupations. To verify that the pretrained LMs evaluated in this work contain background knowledge mapping occupations to situations, we ran a LAMA probe \citep{petroni2020how} on BERT and ELMo. Given the template “The work of a [MASK] is [SITUATION].”, we compared the probabilities the LMs assigned to all single-token occupation names used in KITMUS (probing for multi-token words is not supported by LAMA). BERT assigned higher probabilities to the correct occupation than to any other occupation for 90\% of occupations. ELMo assigned the highest probability to the correct occupation for only 45\% occupations, which might contribute to explaining why the ELMo-based model C2F generally performs worse than BERT4Coref on the {\bgparam} variant KITMUS, which requires such knowledge about occupations.

\para{Root Word Overlap:} One potential limitation of testing for non-fictional background knowledge like \enquote{firefighters put out fires} is that the natural occurrence of the root word \enquote{fire} in both occupation and situation might enable models to solve the task without having access to background knowledge. An analysis of trigram overlaps in all occupation-situation pairs shows that 45\% of non-fictional occupations have at least one overlapping root word. However, a comparison of performances on those samples with and without root word overlap showed neither systematic increase nor decrease for any model, indicating that models do not rely on the root word mappings. Results split up by root word overlap can be found in Table \ref{tab:bgparam-overlap}.

\para{Train Set Size:}
The size of the train set for {\Dataname}, 2000, was chosen to mirror that of GAP \citep{webster-etal-2018-mind}. To evaluate whether the failure of models to learn the task is due to the relatively small number of samples observed during training, we re-generated all variants with 5000 train examples and repeated all experiments. We observe an increase in the magnitude of performance both in BERT4Coref and C2F on those variants where performance was higher than random performance with 2000 examples, but not on those that were equal to or below random performance. Consistent with previous results, BERT4Coref performs well on {\bgparam} and {\bgparamcont}, but not on all fictional {\bgcont} variants (Tables~\ref{tab:bgparam-5k} and \ref{tab:bgothers-5k}). We release the {\Dataname} generation code to enable experimentation with other train set sizes in future work.

\section*{Ethical Considerations}
\label{sec:ethical-considerations}

While {\Dataname} is intended as a diagnostic tool, users should be aware of the possibility of unintended biases when interpreting model performances on this dataset. To document these in more detail, our dataset release will be accompanied by a datasheet \citep{gebru2018datasheets} which is included in Appendix \ref{sec:appendix-datasheet}.

Despite the synthetic nature, depending on its use, {\Dataname} might also have adverse impacts. The randomized sampling of resources to fill slots is meant to minimize bias in terms of the demographic cues that might be associated with the entities referenced in our tests (e.g., gender and nationality). The names and occupation descriptions in our test suite are drawn from United States governmental resources or English-language websites. This means that our test suite is not representative and likely skewed in terms of names, locations, occupations, and situations more common in the e.g., anglophone world. Additional resources such as noise statements and fictional entities were generated using word-level and character-level language models trained on English-language texts, which are known to reproduce a variety of biases found in natural data \citep{bordia-bowman-2019-identifying, solaiman2019release}.

Our human validation study was IRB approved.

\section*{Acknowledgements}
\label{sec:acknowledgements}

We would like to thank the anonymous reviewers for their valuable suggestions. This work was supported by Microsoft Research. Jackie Chi Kit Cheung is supported by the Canada CIFAR AI Chair program, and is also a consulting researcher for Microsoft Research. The authors acknowledge the material support of NVIDIA in the form of computational resources. This research was enabled in part by compute resources provided by Mila (\url{mila.quebec}).

\bibliography{references/anthology, references/custom}
\bibliographystyle{acl/acl_natbib}

\appendix

\section{Appendix}

\subsection{Creating a Custom Dataset}
\label{sec:custom-dataset-creation}

Our code can be used to create a custom dataset in different languages by using custom resources in place of the canonical resources listed in \ref{sec:canonical-resources}.

Detailed instructions for how to do this can be found in the code repository's README\footnote{\url{https://github.com/mpoemsl/kitmus/blob/main/README.md}} file.

\subsection{Dataset-specific Resources}
\label{sec:canonical-resources}

This section details the resources that were used to create the {\Dataname} dataset.

\subsubsection{Templates}
\label{sec:templates}

Table \ref{tab:templates} shows the sets of templates used to to introduce and refer to entities.

\begin{table*}[t!]
    \footnotesize
    
    \begin{subtable}[h]{0.40\textwidth}
        At \{location\}, \{mentions\} met.\\
        At \{location\}, \{mentions\} ran into each other.\\
        At \{location\}, \{mentions\} started a conversation.\\
        At \{location\}, \{mentions\} came across each other.\\
        At \{location\}, \{mentions\} encountered each other.\\
        At \{location\}, \{mentions\} bumped into each other.\\
        At \{location\}, \{mentions\} connected.\\
        \{mentions\} met at \{location\}.\\
        \{mentions\} ran into each other at \{location\}.\\
        \{mentions\} started a conversation at \{location\}.\\
        \{mentions\} came across each other at \{location\}.\\
        \{mentions\} encountered each other at \{location\}.\\
        \{mentions\} bumped into each other at \{location\}.\\
        \{mentions\} connected at \{location\}.\\
        \vspace{-2ex}
        \caption{Meet Sentence Templates}
    \end{subtable}
    \hfill
    \begin{subtable}[h]{0.55\textwidth}
        After a long day at work \{situation\}, \{pronoun\} was happy to relax.\\
        \{pronoun\} told anecdotes from a career of \{situation\}.\\
        \{pronoun\} reflected on whether \{situation\} for a living was a good career choice.\\
        When a question related to \{situation\} arose, \{pronoun\} offered a professional opinion.\\
        \{pronoun\} was relieved to unwind after a demanding day at work \{situation\}.\\
        \{pronoun\} was glad to unwind after a long day at work \{situation\}.\\
        \{pronoun\} shared experiences from a career of \{situation\}.\\
        \{pronoun\} pondered whether choosing \{situation\} as a career was a wise decision.\\
        \vspace{3ex}
        \caption{Pronoun Sentence Templates}
    \end{subtable}
    
\vspace{-6pt}
\caption{Templates used to introduce (\enquote{Meet Sentence}) and refer to (\enquote{Pronoun Sentence}) entities in {\Dataname} task.}
\label{tab:templates}
\end{table*}

\subsubsection{Fictional Occupations and Situations}
\label{sec:appendix-fictional-occupation}

We generally follow the methodology of \citet{malkin-etal-2021-gpt} in creating fictional occupations and siutations. To bias the model towards strings that can be used as occupation names, we train it on a reversed sequence of characters and prompt with the suffix \texttt{er}. We manually filter the words to eliminate unpronounceable or pre-existing English words.

We employ the following two methodologies to generate fictional situations: 1) character-level fictional---like the fictional occupations---is generated with the suffix prompts \texttt{ing} and \texttt{ly}, and 2) word-level fictional is generated by randomly shuffling existing words with the same POS tags followed by manual filtering based on semantic plausibility. Examples are shown in Table \ref{table:occupation-comb}.

\subsubsection{Resource Pools}
\label{sec:appendix-resource-pool}

\textbf{Entities} are sampled from a pool of the $20{,}000$ most frequent last names in the 2010 U.S. census.\footnote{\url{https://www.census.gov/topics/population/genealogy/data/2010_surnames.html}} We use last names as entity names in order to avoid introducing gender-related cues. We discard those last names that are also first names. The order of entities within a template is also randomized. We assume that there is no confounding pretrain-time knowledge based on the entity names in the models.

\textbf{Occupations} consist of a curated list of $60$ common occupations compiled by scraping a career website\footnote{\url{https://ca.indeed.com/career-advice/finding-a-job/common-jobs}} and the US Labor census data.\footnote{\url{https://www.bls.gov/emp/tables/emp-by-detailed-occupation.htm}} Following \citet{cao-daume-iii-2020-toward}, we remove referential gender cues from the occupations such as \enquote{fireman.} The jobs pertaining to very specific domains or related to one of the locations where entities meet are removed from the list.

\textbf{Situations} are assembled using the occupation descriptions of the scraped occupations. We manually filter the pairs of situations that are semantically similar, such as an accountant and an analyst.

\textbf{Locations} are derived from a curated list of $112$ locations scraped from a website of common meet-up places.\footnote{\url{https://www.happierhuman.com/meet-new-people/}} We manually filter out locations that could provide inadvertent surface cues related to the entities' occupation, nationality, or gender.

\subsection{Dataset Format}
\label{sec:appendix-data-format}

The CoNLL format contains token and sentence boundaries, Penn Treebank POS tags \citep{marcinkiewicz1994building}, and gold coreference clusters for all entity mentions. This means that all mentions of an entity---including in the knowledge text---are annotated in a single cluster. Models that operate on the CoNLL format predict these clusters, which involves both detecting mentions and clustering them. In contrast, the GAP format allows for the annotation of only two entities and only one mention per entity (excluding the pronoun), so entity mentions in the knowledge text remain un-annotated. Models that operate on the GAP format are presented with exactly two mentions and for each of them make a binary decision whether or not they are coreferring with a pronoun. The GAP format task is more restricted in that models do not have to detect mentions and there are at most two entities per instance.

\subsection{Human Validation}
\label{sec:appendix-human-val}

The participants were undergraduate and graduate students with fluency in English which were recruited via an institution-wide open call. The participants were compensated with the equivalent of $12$ USD for their participation.\footnote{Matches the minimum wage in the participants' demographic} The study was approved by the institution's ethics review board and the participants gave their written consent via a form.

The participants were tasked to resolve the coreferences in a randomly sampled subset of {\Dataname} texts. The task is presented to the participants as a multiple choice questionnaire. The participants are given gold mentions and have to select the antecedent that is referred to by the pronoun. The answer options include the names of all mentioned entities and a \enquote{can't say} option to indicate that the question is not answerable. The questionnaire contained $96$ questions to be completed in $60$ minutes, which was generous for most participants. 

\begin{figure*}[t!]
    \footnotesize
    
    \begin{subfigure}[h]{0.45\textwidth}
        \centering
        \includegraphics[width=\linewidth]{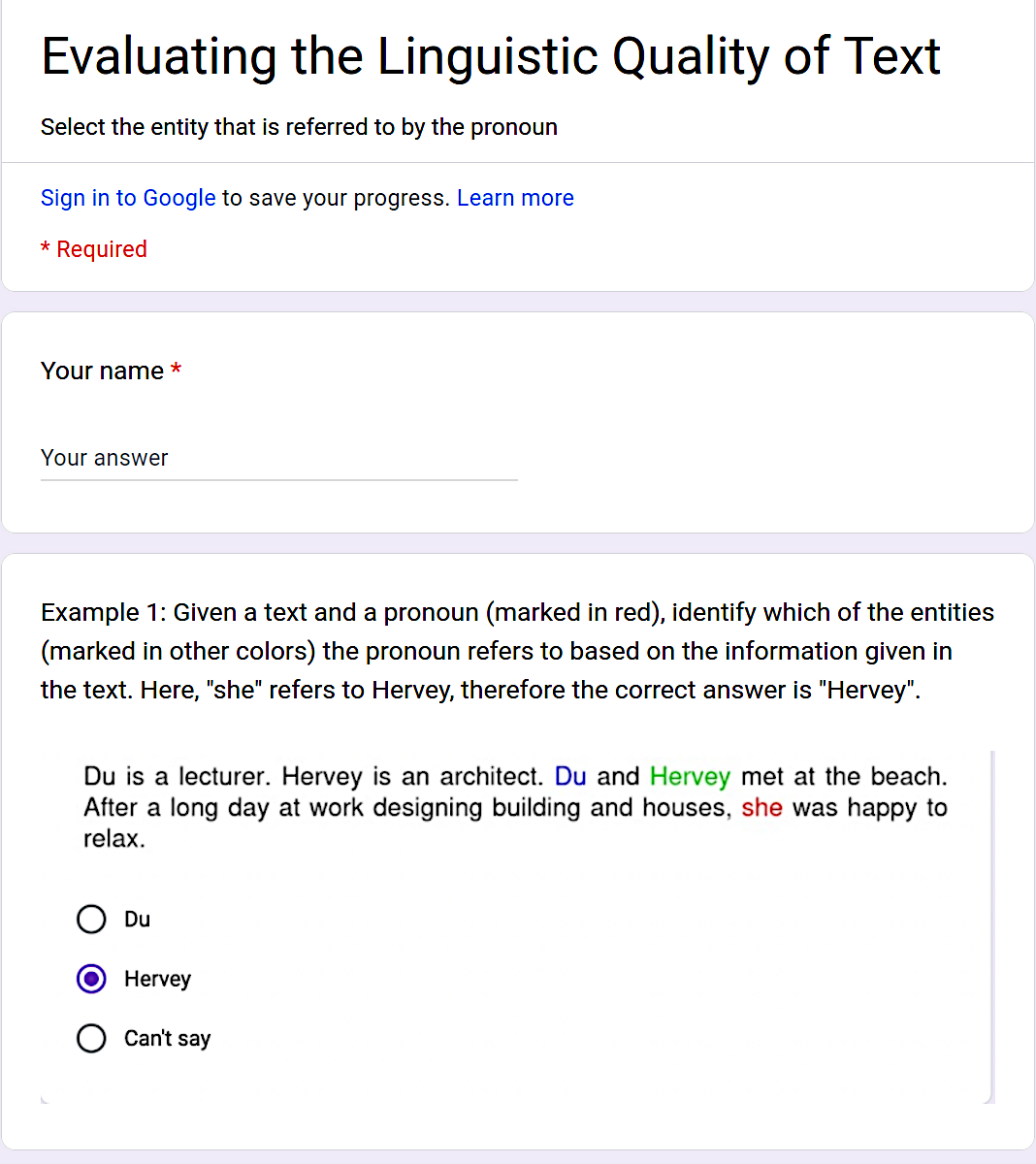}
        \caption{Top Half}
    \end{subfigure}
    \hfill
    \begin{subfigure}[h]{0.45\textwidth}
        \centering
        \includegraphics[width=\linewidth]{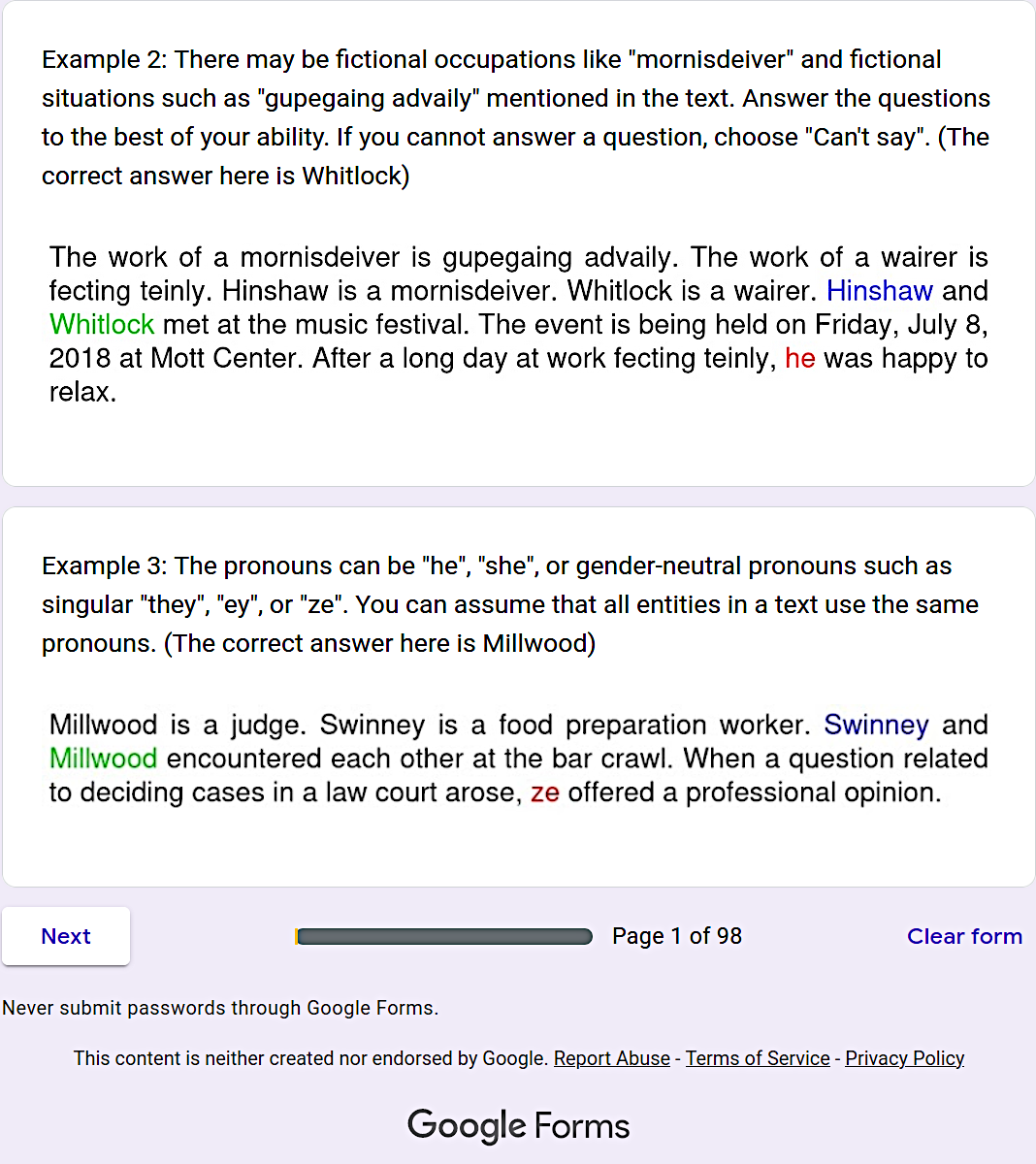}
        \caption{Bottom Half}
    \end{subfigure}
    
    \vspace{-8pt}
    \caption{Human validation questionaire introduction (split into two halves because of space constraints).}
    \label{fig:human-val-form}
\end{figure*}

The human validation was conducted using Google forms. The participants are introduced to the task with examples as shown in Figure~\ref{fig:human-val-form}.

This is followed by 96 questions where the participants have to choose one option among all entity names and the option \enquote{can't say,} which indicates that the task cannot be solved for this instance. The aggregated results of the validation study are shown in Table \ref{tab:human-val-results}.

\begin{table*}[t!]
    \centering \footnotesize
    \begin{tabular}{lllcc} 
    Variant & Occupation & Situation & With Knowledge & Without Knowledge \\
    \toprule

    {\bgparam} & \multirow{2}{*}{Real} & \multirow{2}{*}{Real} & 0.93 & 0.00 \\
    {\bgparam} without noise & & & 0.91 & 0.00 \\
    
    \midrule

    {\bgparamcont} & \multirow{3}{*}{Real} & Real & 1.00 & 0.00 \\
    {\bgcont} &  & CharFict & 1.00 & 0.00 \\
    {\bgcont} &  & WordFict & 0.98 & 0.00 \\

    \midrule

    {\bgcont} & \multirow{3}{*}{CharFict} & Real & 0.98 & 0.00 \\
    {\bgcont} &  & CharFict & 0.98 & 0.00 \\
    {\bgcont} &  & WordFict & 0.96 & 0.06 \\
    
    \bottomrule
    
    \end{tabular}
    \vspace{-8pt}
    \caption{Accuracy on all variants aggregated over subtasks, splits, and participants. Random performance is $0.25$. Human participants could select \enquote{can't say,} which is never in agreement with the automatically generated labels.}
    \label{tab:human-val-results}
\end{table*}

\subsection{Training Details}
\label{sec:appendix-train-time}

We train our models in a compute cluster infrastructure on Nvidia Quadro RTX 8000 GPUs. For BERT4Coref, training on the train split of one {\Dataname} subtask took about $8$ hours per run. For C2F it took about $16$ hours, the training of the ensemble model GREP took $18$ hours. The training of smaller models and inference on pretrained models took about $4$ hours per run.

\subsection{Additional Experiments}
\label{sec:appendix-add-exps}

As a supplement to our main experiments, we report the following experiment results on the {\bgparam} variant:
\begin{compactitem}
    \item F1 score in Table \ref{tab:bgparam-f1}
    \item Accuracy with 5000 instead of 2000 train examples in Table \ref{tab:bgparam-5k}
    \item Accuracy without noise in Table \ref{tab:bgparam-noise}
    \item Accuracy on train set in Table \ref{tab:bgparam-train}
    \item Accuracy with and without root word overlap in Table \ref{tab:bgparam-overlap}
\end{compactitem}

On the {\bgparamcont} and {\bgcont} variants, we report:
\begin{compactitem}
    \item F1 score in Table \ref{tab:bgothers-f1}
    \item Accuracy on train set in Table \ref{tab:bgothers-train}
    \item Accuracy with 5000 instead of 2000 train examples in Table \ref{tab:bgothers-5k}
\end{compactitem}

\begin{table*}[t!]
    \footnotesize
    
    \begin{subtable}[h]{0.45\textwidth}
        \centering
        \begin{tabular}{lccc}
        Model & 2 Entities & 3 Entities & 4 Entities \\
        
        \midrule
    
        BERT4Coref & \textbf{0.49} & 0.24 & 0.19 \\
        C2F & 0.48 & \textbf{0.33} & \textbf{0.25} \\

        Stanford Neural & 0.29 & 0.15 & 0.13 \\
        Stanford Stat. & 0.09 & 0.04 & 0.02 \\
        
        \midrule
        Random & 0.50 & 0.33 & 0.25 \\
        
        \bottomrule
        \end{tabular}
        \caption{Ontonotes-trained}
    \end{subtable}
    \hfill
    \begin{subtable}[h]{0.45\textwidth}
        \centering
        \begin{tabular}{lccc}
        Model & 2 Entities & 3 Entities & 4 Entities \\
        
        \midrule
        
        BERT4Coref & \textbf{0.99} & \textbf{0.99} & \textbf{0.94} \\
        C2F & 0.52 & 0.35 & 0.48 \\
        
        GREP\textsuperscript{\textdagger} & 0.49 & - & -\\
        PeTra\textsuperscript{\textdagger} & 0.67 & - & -\\
        
        \midrule
        Random & 0.50 & 0.33 & 0.25 \\
        
        \bottomrule
        \end{tabular}
        \caption{{\Dataname}-trained}
    \end{subtable}
    
    \vspace{-8pt}
    \caption{Antecedent F1 on {\bgparam} variant of {\Dataname}. Models marked with {\textdagger} operate on GAP format which only allows for the annotation of two entities. All other models operate on the CoNLL format. PeTra has higher F1 scores than pronoun accuracy, since it defaults to always predicting \texttt{true} for each antecedent, which results in a recall of $1.00$ and a thus a high F1 score.}
    \label{tab:bgparam-f1}
\end{table*}

\begin{table*}[h!]
    \centering \footnotesize
    \begin{tabular}{lccccccc} 
    \toprule
    
    & & \multicolumn{2}{c}{2 Entities} & \multicolumn{2}{c}{3 Entities} & \multicolumn{2}{c}{4 Entities} \\
    Model & Train Data & 2k & 5k & 2k & 5k & 2k & 5k \\
    
    \midrule
    
    PeTra & \multirow{4}{*}{{\Dataname}} & 0.00 & 0.01 & - & - & - & - \\
    GREP &  & 0.49 & 0.50 & - & - & - & - \\
    BERT4Coref &  & 0.99 $\pm$ 0.00 & 1.00 $\pm$ 0.00 & 0.98 $\pm$ 0.01 & 0.97 $\pm$ 0.00 & 0.94 $\pm$ 0.01 & 0.94 $\pm$ 0.02 \\
    C2F &   & 0.52 $\pm$ 0.02 & 0.58 $\pm$ 0.06 & 0.28 $\pm$ 0.08 & 0.63 $\pm$ 0.03 & 0.48 $\pm$ 0.06 & 0.24 $\pm$ 0.08\\
        
    \midrule
    
    Random & - & 0.50 & 0.50 & 0.33 & 0.33 & 0.25 & 0.25 \\ 
    
    \bottomrule
    \end{tabular}
    
    \caption{Accuracy on {\bgparam} variant of {\Dataname} with $2000$ (2k) and $5000$ (5k) train examples. Standard deviation is given after $\pm$.}
    \label{tab:bgparam-5k}
\end{table*}

\begin{table*}[h!]
    \centering \footnotesize
    \begin{tabular}{lccccccc} 
    \toprule
    
    & & \multicolumn{2}{c}{2 Entities} & \multicolumn{2}{c}{3 Entities} & \multicolumn{2}{c}{4 Entities} \\
    Model & Train Data & Noise & No Noise & Noise & No Noise & Noise & No Noise \\

    \midrule
    
    BERT4Coref & \multirow{4}{*}{Ontonotes} & 0.43 & 0.43 & 0.18 & 0.23 & 0.14 & 0.13 \\
    C2F &  & 0.34 & 0.34 & 0.18 & 0.18 & 0.13 & 0.14 \\
    Stfd. Neural &  & 0.20 & 0.33 & 0.10 & 0.15 & 0.09 & 0.14 \\
    Stfd. Stat. &  & 0.05 & 0.15 & 0.02 & 0.06 & 0.01 & 0.06 \\
    
    \midrule
    
    PeTra & \multirow{4}{*}{{\Dataname}} & 0.00 & 0.01 & - & - & - & - \\
    GREP &  & 0.49 & 0.49 & - & - & - & - \\
    BERT4Coref &  & 0.99 & 1.00 & 0.98 & 0.98 & 0.94 & 0.92 \\
    C2F &   & 0.52 & 0.52 & 0.28 & 0.34 & 0.48 & 0.24 \\
        
    \midrule
    
    Random & - & 0.50 & 0.50 & 0.33 & 0.33 & 0.25 & 0.25 \\ 
    
    \bottomrule
    \end{tabular}
    
    \caption{Accuracy on {\bgparam} variant of {\Dataname} with and without noise.}
    \label{tab:bgparam-noise}
\end{table*}

\begin{table*}[ht!]
    \centering \footnotesize
    \begin{tabular}{lccccccc} 
    \toprule
    
    & & \multicolumn{2}{c}{2 Entities} & \multicolumn{2}{c}{3 Entities} & \multicolumn{2}{c}{4 Entities} \\
    Model & Train Data & Test & Train & Test & Train & Test & Train \\
    
    \midrule
    
    PeTra & \multirow{4}{*}{{\Dataname}} & 0.00 & 0.01 & - & - & - & - \\
    GREP &  & 0.49 & 0.51 & - & - & - & - \\
    BERT4Coref &  & 0.99 & 1.00 & 0.98 & 1.00 & 0.94 & 1.00 \\
    C2F &   & 0.52 & 0.96 & 0.28 & 1.00 & 0.48 & 1.00 \\
        
    \midrule
    
    Random & - & 0.50 & 0.50 & 0.33 & 0.33 & 0.25 & 0.25 \\ 
    
    \bottomrule
    \end{tabular}
    
    \caption{Test and train accuracy on {\bgparam} variant of {\Dataname}.}
    \label{tab:bgparam-train}
\end{table*}

\begin{table*}[ht!]
    \centering \footnotesize
    \begin{tabular}{lccccccc} 
    \toprule
    
    & & \multicolumn{2}{c}{2 Entities} & \multicolumn{2}{c}{3 Entities} & \multicolumn{2}{c}{4 Entities} \\
    Model & Train Data & Overlap & No Overlap & Overlap & No Overlap & Overlap & No Overlap \\

    \midrule
    
    BERT4Coref & \multirow{4}{*}{Ontonotes} & 0.43 & 0.45 & 0.18 & 0.19 & 0.15 & 0.14 \\
    C2F &  & 0.34 & 0.36 & 0.17 & 0.19 & 0.13 & 0.12 \\
    Stfd. Neural &  & 0.20 & 0.19 & 0.11 & 0.08 & 0.08 & 0.09 \\
    Stfd. Stat. &  & 0.05 & 0.04 & 0.02 & 0.01 & 0.01 & 0.00 \\
    
    \midrule
    
    PeTra & \multirow{4}{*}{{\Dataname}} & 0.00 & 0.01 & - & - & - & - \\
    GREP &  & 0.47 & 0.52 & - & - & - & - \\
    BERT4Coref &  & 0.99 & 0.99 & 0.99 & 0.97 & 0.95 & 0.92 \\
    C2F &   & 0.53 & 0.50 & 0.29 & 0.26 & 0.49 & 0.46 \\
        
    \midrule
    
    Random & - & 0.50 & 0.50 & 0.33 & 0.33 & 0.25 & 0.25 \\ 
    
    \bottomrule
    \end{tabular}
    
    \caption{Accuracy on {\bgparam} variant of {\Dataname} with and without root word overlap.}
    \label{tab:bgparam-overlap}
\end{table*}


\begin{table}[ht!]
    \centering \footnotesize
    \begin{tabular}{lllcc} 
    Var. & Occupation & Situation & C2F & BERT4Coref \\
    \toprule
    
    BB & \multirow{3}{*}{Real} & Real &  0.11 & 0.96 \\
     
    BI &  & CharFict & 0.20 & 0.25 \\ 
    BI &  & WordFict & 0.10 & 0.49 \\
     
    \midrule
     
    BI & \multirow{3}{*}{CharFict} & Real & 0.09 & 0.43 \\
    BI & & CharFict & 0.21 & 0.27 \\ 
    BI & & WordFict & 0.14 & 0.39 \\
    \bottomrule
    
    \end{tabular}
    \vspace{-8pt}
    \caption{{\Dataname}-trained F1 Score on {\bgparamcont} (BB) and {\bgcont} (BI) variants of {\Dataname} with four entities. Random performance is 0.25.}
    \label{tab:bgothers-f1}
\end{table}

\begin{table*}[t!]
    \centering \footnotesize
    \begin{tabular}{lllcccc} 

     &  &  & \multicolumn{2}{c}{C2F} & \multicolumn{2}{c}{BERT4Coref} \\

    Variant & Occupation & Situation & Test & Train & Test & Train \\
    \toprule
    
    BB & \multirow{3}{*}{Real} & Real & 0.09 & 1.00 & 0.96 & 1.00 \\
     
    BI &  & CharFict & 0.18 & 0.97 & 0.25 & 0.88 \\ 
    BI &  & WordFict & 0.08 & 0.95 & 0.48 & 0.73 \\
     
    \midrule
     
    BI & \multirow{3}{*}{CharFict} & Real & 0.08 & 0.96 & 0.43 & 0.97 \\
    BI & & CharFict & 0.18 & 0.83 & 0.26 & 0.78 \\ 
    BI & & WordFict & 0.11 & 1.00 & 0.38 & 0.96 \\

    \bottomrule
    
    \end{tabular}
    \vspace{-8pt}
    \caption{Train and test accuracy on {\bgparamcont} (BB) and {\bgcont} (BI) variants of {\Dataname}. Random performance is $0.25$.}
    \label{tab:bgothers-train}
\end{table*}

\begin{table*}[t!]
    \centering \footnotesize
    \begin{tabular}{lllcccc} 

     &  &  & \multicolumn{2}{c}{C2F} & \multicolumn{2}{c}{BERT4Coref} \\

    Variant & Occupation & Situation & 2k & 5k & 2k & 5k \\
    \toprule
    
    BB & \multirow{3}{*}{Real} & Real & 0.09 & 0.49 & 0.96 & 0.97 \\
     
    BI &  & CharFict & 0.18 & 0.25 & 0.25 & 0.27 \\ 
    BI &  & WordFict & 0.08 & 0.26 & 0.48 & 0.78 \\
     
    \midrule
     
    BI & \multirow{3}{*}{CharFict} & Real & 0.08 & 0.21 & 0.43 & 0.57 \\
    BI & & CharFict & 0.18 & 0.25 & 0.26 & 0.26 \\ 
    BI & & WordFict & 0.11 & 0.25 & 0.38 & 0.59 \\
    \bottomrule
    
    \end{tabular}
    \vspace{-8pt}
    \caption{{\Dataname}-trained accuracy on {\bgparamcont} (BB) and {\bgcont} (BI) variants of {\Dataname} with four entities with $2000$ (2k) and $5000$ (5k) train examples. Random performance is $0.25$.
    }
    \label{tab:bgothers-5k}
\end{table*}


\subsection{Datasheet}
\label{sec:appendix-datasheet}

\subsubsection{Motivation}

\textbf{For what purpose was the dataset created?}

The {\Dataname} dataset was created to enable research on reasoning over knowledge for the task of coreference resolution - i.e. given a piece of text, identify mentions and determine whether or not they co-refer. The dataset was created with the intention to focus on those cases of coreference resolution that require knowledge about specific entities and their occupations to accomplish the task.

\textbf{Who created the dataset and on behalf of which entities?}

The dataset was created by the authors of this paper.

\textbf{Who funded the creation of the dataset?}

Funding was provided by multiple sources as mentioned in the acknowledgements in section \ref{sec:acknowledgements}.

\textbf{Any other comments?}

None.

\subsubsection{Composition}
\label{subsubsec:composition}

\textbf{What do instances that comprise the dataset represent?}

The dataset consist of text pairs that were generated to capture knowledge about entities, occupations, and situations, as well as coreference cases whose resolution depends on this knowledge. The labels are clusters of tokens in the text.

\textbf{How many instances are there in total?}

There are $4400 \cdot 3 \cdot (2 + 1 + 5) = 105600$ instances in total: $4400$ instances for each of the three entity numbers for variants {\bgparam} (also without noise), {\bgparamcont}, and five versions of {\bgcont} with different degrees of fictionality.

\textbf{Does the dataset contain all possible instances or is it a sample of instances from a larger set?}

The dataset contains all instances that we generated. They are generated by filling slots in a template by sampling from a pool of resources. The pool of resources only contains a subset of resources in the world, and the sampling process selects a random subset of the pool of resources.

\textbf{What data does each instance consist of?}

The instances are pairs of template-generated texts: one knowledge text and one task text. The knowledge text contains knowledge about fictional entities and real or fictional occupations in text form. The task text contains a case of coreference involving the same fictional entities. Labels for the coreferences are given in the form of coreference clusters over tokens.

\textbf{Is there a label associated with each instance?}

Yes. The label is a coreference cluster that represents the true resolution of the coreference presented in the text.

\textbf{Is any information missing from individual instances?}

No.

\textbf{Are relationships between individual instances made explicit?}

Yes. The entities are fictional and created separately for each instance. Instances are completely independent from each other and are not consistent across the dataset, i.e. conflicting knowledge may be given for the same fictional entity across different instances in the dataset.

\textbf{Are there recommended data splits?}

Yes. Each subcategory of the dataset is provided in recommended data splits of $2000$ \texttt{train} instances, $400$ \texttt{validation} instances, and $2000$ \texttt{test} instances. The numbers are chosen for size comparability with other coreference resolution datasets such as GAP \citep{webster-etal-2018-mind}. Resources are disjunct across the splits for each subcategory, which enables the evaluation of the ability of models to generalize beyond observed resources.

\textbf{Are there any errors, sources of noise, or redundancies in the dataset?}

None that we are aware of. Since the dataset is template-generated, only the intentionally provided noise in the appropriate subcategory is present. We control for redundancies in the dataset. A human validation has not brought to light any errors in the dataset, however, due to the synthetic nature of the dataset texts can appear wooden and non-natural to readers.

\textbf{Is the dataset self-contained, or does it link to or otherwise rely on external resources?}

The dataset is created using external resources to fill slots in templates, but the finished dataset is entirely self-contained.

\textbf{Does the dataset contain data that might be considered confidential?}

The dataset contains only information about fictional entities and public knowledge about occupations which is not confidential.

\textbf{Does the dataset contain data that, if viewed directly, might be offensive, insulting, threatening, or might otherwise cause anxiety?}

Both the templates and the resources used to fill the slots were manually inspected for content that might cause anxiety to viewers. 

The dataset does not contain any text that might cause anxiety to viewers.

\textbf{Does the dataset identify any subpopulations?}

The fictional entities have neither an explicit age nor gender. The only distinguishing features of the entities are their names and occupations, which are uniformly sampled, and their pronoun use, which is sampled according to the following distribution: $40\%$ \texttt{he}, $40\%$ \texttt{she}, $10\%$ \texttt{they}, and $10\%$ neopronouns.

\textbf{Is it possible to identify individuals either directly or indirectly?}

No. Since the entities are entirely fictional, any similarities to existing individuals are due to chance.

\textbf{Does the dataset contain data that might be sensitive in any way?}

No.

\textbf{Any other comments?}

None.

\subsubsection{Collection Process}

\textbf{How was the data associated with each instance acquired?}

The data was generated by filling slots in templates that were hand-engineered. The slot-filling resources were obtained from publicly available raw text sources such as governmental name statistics and professional job websites. Noise sentences were generated with the language model GPT-2 \citep{radford2019language} and manually edited and verified to conform with the rest of the dataset. Fictional occupation names and descriptions were created by random sampling from a character-level LSTM language model following methodology of \citet{malkin-etal-2021-gpt}.

\textbf{What mechanisms or procedures were used to collect the data?}

The dataset was generated using Python scripts, which will be made publicly available in a GitHub repository.

\textbf{If the dataset is a sample from a larger set, what was the sampling strategy?}

Not applicable. The entire dataset will be released.

\textbf{Who was involved in the data collection process and how were they compensated?}

Not applicable. There was no human involved in the dataset creation prcoess.

\textbf{Over what timeframe was the data collected?}

The dataset was created immediately prior to the submission of this draft for review.

\textbf{Were any ethical review processes conducted for the data collection process?}

Not applicable, data was not collected. The human evaluation study used to evaluate the dataset was approved by an institutional review board.

\textbf{Did you collect the data from the individuals in question directly, or obtain it via third parties or other sources?}

The dataset was created via templates. The resources were collected directly from publicly available data online.

\textbf{Were the individuals in question notified about the data collection?}

The resources were collected directly online from institutions and authors who made the resources available publicly. The authors and institutions were not explicitly informed about the way their resources are used in this dataset.

\textbf{Did the individuals in question consent to the collection and use of
their data?}

Not applicable.

\textbf{If consent was obtained, were the consenting individuals provided with a mechanism to revoke their consent in the future or for certain uses?}

Not applicable.

\textbf{Has an analysis of the potential impact of the dataset and its use on data subjects been conducted?}

No.

\textbf{Any other comments?}

None.

\subsubsection{Preprocessing}

\textbf{Was any preprocessing/cleaning/labeling of the data done?}

The template building blocks were manually tokenized and POS tagged with the Stanford CoreNLP pipeline, which was then manually verified. In terms of resources, the occupations were filtered manually to avoid overlaps in descriptions. Referential gender cues such as \enquote{fireman} were removed from the occupations. Occupations pertaining to very specific domains or related to location were removed from the list. GPT-2 generated noise sentences were manually checked for coherence and also tokenized and POS tagged with the Stanford CoreNLP pipeline. Fictional occupation names and descriptions were likewise manually checked for coherence and suitability.

\textbf{Was the \enquote{raw} data saved in addition to the preprocessed/cleaned/labeled data?}

No.

\textbf{Is the software that was used to preprocess/clean/label the data available?}

The Stanford CoreNLP pipeline is available here: \url{https://stanfordnlp.github.io/CoreNLP/}.

\textbf{Any other comments?}

None.

\subsubsection{Uses}

\textbf{Has the dataset been used for any tasks already?}

None.

\textbf{Is there a repository that links to any or all papers or systems that use the dataset?}

Not applicable.

\textbf{What (other) tasks could the dataset be used for?}

The dataset could potentially be used for research on mention detection, cross-document coreference resolution, or entity linking, since the annotations are compatible with these tasks as well.

\textbf{Is there anything about the composition of the dataset or the way it was collected and preprocessed/cleaned/labeled that might impact future uses?}

Due to its template-generated nature, the data does not consist of naturally occurring texts and should not be used for purposes which require naturally occurring texts.

\textbf{Are there tasks for which the dataset should not be used?}

The entities in the texts are entirely fictional and have an arbitrary distribution of attributes. Consequently, the information in this dataset should not be used to make decisions about real people.

\textbf{Any other comments?}

None.

\subsubsection{Distribution}

\textbf{Will the dataset be distributed to third parties outside of the entity on behalf of which the dataset was created?}

Yes, the dataset will be available publicly on the internet.

\textbf{How will the dataset be distributed?}

The dataset will be released in the GitHub repository for this paper.

\textbf{When will the dataset be distributed?}

Upon publication of the corresponding paper.

\textbf{Will the dataset be distributed under a copyright or other intellectual property (IP) license, and/or under applicable terms of use (ToU)?}

The dataset and the code used to generate it will be distributed under the license specified in the GitHub repository for the dataset. In the repository, we will also request to cite the corresponding paper if the dataset is used.

\textbf{Have any third parties imposed IP-based or other restrictions on the data associated with the instances?}

None that we are aware of.

\textbf{Do any export controls or other regulatory restrictions apply to the dataset or to individual instances?}

None that we are aware of.

\textbf{Any other comments?}

No.

\subsubsection{Maintenance}

\textbf{Who will be supporting/hosting/maintaining the dataset?}

The first authors will support and maintain the dataset.

\textbf{How can the owner/curator/manager of the dataset be contacted?}

Contact the first authors.

\textbf{Is there an erratum?}

No. Future updates and known errors will be specified in the \texttt{README.md} of the repository.

\textbf{Will the dataset be updated?}

Currently, no updates are planned.

\textbf{If the dataset relates to people, are there applicable limits on the retention of the data associated with the instances?}

Not applicable, since the entities are fictional.

\textbf{Will older versions of the dataset continue to be supported/hosted/maintained?}

In the case of updates, the original version of the dataset will always be available on GitHub via a tagged release.

\textbf{If others want to extend/augment/build on/contribute to the dataset, is there a mechanism for them to do so?}

Suggestions for the augmentation of the dataset can be made via GitHub pull requests.

\textbf{Any other comments?}

None.

\end{document}